%% file: main_arxiv.tex
\documentclass{amsart}
\usepackage{amsaddr}

\usepackage[utf8]{inputenc}
\usepackage{amssymb}
\usepackage{amsthm}
\usepackage{amsfonts}
\usepackage{amsmath}
\usepackage{multicol,multirow}
\usepackage[text={440pt,575pt},headheight=9pt,centering]{geometry}

\usepackage{amsmath}
\usepackage{amssymb}
\usepackage{amsthm}
\usepackage{soul}
\usepackage{cancel}
\usepackage{bm}

\usepackage{graphicx}
\usepackage{graphicx}
\usepackage{subcaption}
\usepackage{booktabs}
\usepackage[colorlinks,citecolor=blue,urlcolor=blue]{hyperref}
\usepackage{algorithm}
\usepackage{algpseudocode}

\usepackage{xcolor}
\definecolor{bronze}{RGB}{205, 127, 50}
\definecolor{silver}{RGB}{205, 127, 50}
\definecolor{gold}{RGB}{205, 127, 50}

\usepackage{fullpage}
\usepackage{enumitem}

\usepackage[numbers,sort&compress]{natbib}

\usepackage[normalem]{ulem}
\usepackage{placeins}

\title{Chunked TabPFN: Exact Training-Free In-Context Learning for Long-Context Tabular Data}
\author{Renat Sergazinov$^{*}$}
\address{Department of Statistics, Texas A\&M University, College Station, TX}

\author{Shao-An Yin$^{*}$}
\address{Department of Electrical and Computer Engineering, University of Minnesota, Twin City, MN}
\thanks{$^*$ Equal contribution.}

\begin{document}

\begin{abstract}
TabPFN v2 achieves better results than tree-based models on several tabular benchmarks, which is notable since tree-based models are usually the strongest choice for tabular data. However, it cannot handle more than 10K context tokens because transformers have quadratic computation and memory costs.

Unlike existing approaches that rely on context compression, such as selecting representative samples via K-nearest neighbors (KNN), we introduce a \textbf{tiled-block} strategy to compute attention within the TabPFN framework. This design is compatible with standard GPU setups and, to the best of our knowledge, is the first to enable TabPFN to \textbf{process long contexts without any pre-processing}. We demonstrate the effectiveness of our approach on the standard TabArena benchmark, with code available at \href{https://github.com/mrsergazinov/chunk_tabpfn}{\texttt{chunk\_tabpfn}}.
\end{abstract}

\maketitle


\section{Introduction}
Even though deep learning foundation models that capture the underlying data manifold have shown promising results in language and vision by enabling broad transfer without fine-tuning, applying the same techniques to tabular data remains highly challenging. A key difficulty is that deep learning methods often underperform compared to tree-based models on tabular benchmarks. Recent work in top venues has advanced deep tabular learning through architectures and priors specifically tailored for tabular data. Representative milestones include TabNet \citep{arik2021tabnet}, FT-Transformer \citep{gorishniy2021revisiting}, NODE \citep{popov2019neural}, efficient MLP models \citep{gorishniy2024tabm}, retrieval-based architectures such as TabR \citep{gorishniy2023tabr} and ModernNCA \citep{ye2024modern}, as well as zero-shot foundation models such as TabPFN \citep{hollmann2022tabpfn, hollmann2025accurate} and TabDPT \citep{ma2024tabdpt}. Still, most deep tabular models need to be trained for each specific dataset and task, which means they are not true foundation models that can serve as plug-and-play solutions.

Recent advances in \textbf{in-context learning} and foundation models for large language models (LLMs) suggest promising directions for developing tabular foundation models. Inspired by in-context learning, the family of \textbf{Prior-Data Fitted Networks (PFNs)} eliminates the need for per-dataset training, thereby offering a unified foundation model for tabular data. PFNs achieve this by training a single transformer once to approximate the posterior predictive distribution $p(y \mid x, D_{\text{train}})$ on synthetic tasks sampled from a prior. At test time, predictions on a new tabular dataset can then be obtained in a single forward pass without any gradient updates. In this work, we build on the TabPFN v2 implementation \citep{hollmann2022tabpfn, hollmann2025accurate}.

However, the \textbf{limited context window} remains a practical limitation of PFN models. The original TabPFN \citep{hollmann2022tabpfn} could handle datasets with at most 3{,}000 samples, while the later version by \citet{hollmann2025accurate} extended this to 10{,}000 samples. Nevertheless, this scale is still restrictive in many real-world scenarios. To mitigate this issue, several works have proposed pre-processing strategies to reduce the effective context size, which can be broadly categorized into two approaches: \textbf{clustering the training set} \citep{hollmann2025accurate, xu2024mixture, thomas2024retrieval} and \textbf{data compression} \citep{feuer2024tunetables}.

Within the category of clustering-based approaches, \citet{xu2024mixture} proposed Mixture of In-Context Prompters (MICP), which forms a fixed set of in-context prompts, learns a routing mechanism, and further fine-tunes TabPFN using bootstrapping within each cluster. Similarly, \citet{hollmann2025accurate} introduced an ensemble-like method that trains a random forest to partition the data into subsets, on top of which TabPFN is applied. In parallel, \citet{thomas2024retrieval} proposed a $k$NN-based strategy, where the “clusters” are not predefined but instead sampled per test point, at the cost of increased computational overhead.

On the other hand, within the category of data-compression methods, \citet{feuer2024tunetables} proposed \textit{TuneTables}, which learns compact, dataset-specific embeddings to replace or augment the context. TuneTables optimizes a small table of embeddings that serves as a surrogate for the full dataset, thereby improving both accuracy and latency through parameter-efficient updates. In addition, it explores multi-objective tuning (e.g., accuracy–fairness trade-offs) and provides analyses of different context optimization strategies.

Even though the above approaches have demonstrated strong performance, two fundamental issues remain with respect to addressing the long-context challenge:
\begin{enumerate}[label=\arabic*.]
    \item They require per-dataset hyperparameter tuning and fine-tuning, which undermines the principle of in-context learning and the vision of foundation models, where predictions should be made in a single forward pass without task-specific adaptation.
    \item From a theoretical perspective, the core principle of TabPFN—fitting the posterior through the provided context—is violated when the original long context is replaced by clustered or compressed representations.
\end{enumerate}

Given the above fundamental limitations, directly using all training examples as context is preferable from both practical and theoretical perspectives. In this paper, we address the long-context challenge \emph{without} any pre-processing by posing the following research questions:
\begin{enumerate}[label=\arabic*.]
    \item How can the memory complexity of long contexts be resolved without pre-processing the training examples?
    \item How well does TabPFN perform in a direct, one-shot in-context learning setting on long contexts without any pre-processing?
\end{enumerate}

Our contributions are summarized as follows:
\begin{enumerate}[label=\arabic*.]
    \item \textbf{Training-free long-context extension.} We introduce a simple, pure-PyTorch modification that computes attention in chunks, supporting both FlashAttention~\cite{FlashAttention} and native PyTorch scaled dot-product attention. This enables TabPFN to operate on datasets with over 100K samples without any retraining.
    \item \textbf{Empirical analysis.} We systematically evaluate TabPFN~v2 on the TabArena benchmark. On a curated subset of 17 long-context tasks, we observe consistent performance gains from additional in-context examples, even beyond the original pre-training limits. This establishes a strong \emph{training-free} baseline for future research and challenges the prevailing assumption that TabPFN models cannot scale beyond their pre-training context window.
\end{enumerate}

\section{Methodology}
In this section, we introduce our chunked TabPFN method. We begin by reviewing the standard attention computation and its role in PFN models. Next, we identify the key memory complexity bottleneck that prevents standard TabPFN from scaling to long contexts. Finally, we present our approach, which partitions the input into chunks and performs attention operations in a batch-wise manner, thereby mitigating out-of-memory (OOM) issues.

\paragraph{\textbf{Notation.}}
We denote the training set (context) as $D_{\text{train}} = \{(x_i, y_i)\}_{i=1}^{n}$, where each feature vector is $x \in \mathbb{R}^p$ with $p$ features and the corresponding label is $y \in \mathcal{Y}$. A test input is denoted by $x_\ast$, and the test set is given by $D_{\text{test}} = \{(x_j, y_j)\}_{j=1}^{m}$.

\paragraph{\textbf{Attention shapes.}}
For the attention block, let $L$ denote the total sequence length consumed by the transformer (i.e., context tokens plus query), $B$ the batch size (number of test points processed in parallel), $H$ the number of attention heads, and $d_k$ the per-head key/query dimension. We adopt the standard notation of Query (Q), Key (K), and Value (V), where the attention operation is applied on
\begin{equation}
\begin{split}
Q &\in \mathbb{R}^{B \times H \times L_q \times d_k}, \\
K &\in \mathbb{R}^{B \times H \times L_k \times d_k}, \\
V &\in \mathbb{R}^{B \times H \times L_k \times d_k},
\end{split}
\end{equation}
with $Q$, $K$, and $V$ being linear projections of the input data sequence. In the common case where $L_q = L_k = L_v$, we simply write $L$ for brevity.

\paragraph{\textbf{PFN model sketch.}}
Prior-Data Fitted Networks (PFNs) train \textbf{a transformer} once on synthetic tasks sampled from a prior, such that at inference time a new tabular dataset can be processed in a single forward pass without gradient updates (i.e., the model parameters remain fixed). PFNs approximate the posterior predictive distribution  
\begin{equation}
p(y_\ast \mid x_\ast, D_{\text{train}}).
\end{equation}
The model ingests $(D_{\text{train}}, x_\ast)$ as a permutation-invariant (set-like) sequence and outputs an approximate predictive distribution  
\begin{equation}
q(y_\ast \mid x_\ast, D_{\text{train}}).
\end{equation}
In the remainder of this work, we focus on the TabPFN implementation \citep{hollmann2022tabpfn,hollmann2025accurate}.

\paragraph{\textbf{Key challenges when scaling context length.}}
According to the above sketch, the core sequence module of TabPFN is multi-head attention. Scaled dot-product attention for a single head is defined as  
\begin{equation}
\mathrm{Attn}(Q,K,V) = \operatorname{softmax}\!\left(\frac{QK^\top}{\sqrt{d_k}}\right)V,
\end{equation}
where the softmax is applied row-wise over the scaled score matrix $QK^\top / \sqrt{d_k}$. The computational and memory costs both scale as $\mathcal{O}(BHL_qL_k)$ due to the size of the score matrix. 

In TabPFN \citep{hollmann2025accurate}, attention is employed in three distinct components:
\begin{enumerate}[label=\arabic*.]
    \item \textbf{Between-feature attention:} $B=n$ and $L=p$.\footnote{TabPFN groups features; thus $L_q$ depends on $p$. We omit this detail for concision.}
    \item \textbf{Self-attention over $D_{\text{train}}$:} $B=p$ and $L=n$ (quadratic in $n$).
    \item \textbf{Cross-attention $D_{\text{train}} \!\to\! D_{\text{test}}$:} $L_q=m$, $L_k=L_v=n$, and $B=p$ (bi-linear in $m$ and $n$).
\end{enumerate}
As the training set size $|D_{\text{train}}| = n$ grows, cases (2) and (3) dominate both runtime and peak memory consumption, even when the architecture and parameters remain fixed.

\paragraph{\textbf{Exact chunking of $Q,K,V$ (framework-native).}}
We compute attention exactly using a \emph{two-level tiling scheme} that chunks all tensors, implemented entirely with stock tensor operations (i.e., framework-native, without custom kernels). Concretely:
\begin{enumerate}[label=\arabic*.]
    \item \textbf{First-level tiling.} $Q$ is split into query tiles of length $\ell$. For each query tile, we iterate over $K/V$ tiles of length $r$ (optionally also tiling along the batch dimension by $m$).
    \item \textbf{Second-level tiling.} For each query position, we maintain a row-wise running maximum $\mu$ and two accumulators: $s$ (the sum of exponentials) and $a$ (the sum of exponentials multiplied by $V$) across all $K/V$ tiles.
    \item \textbf{Final aggregation.} After all tiles have been processed, the output is computed as $a/s$, which is \emph{exactly} equivalent to the monolithic $\operatorname{softmax}(QK^\top) \, V$, up to floating-point associativity.
\end{enumerate}
Peak memory now scales linearly with the tile sizes ($\ell, r$) rather than with the full sequence lengths ($L_q, L_k$), while FLOPs remain unchanged. Further details and pseudocode are provided in Appendix~\ref{app:chunked-attn}.

\begin{figure}[t]
    \centering
    \includegraphics[width=\linewidth]{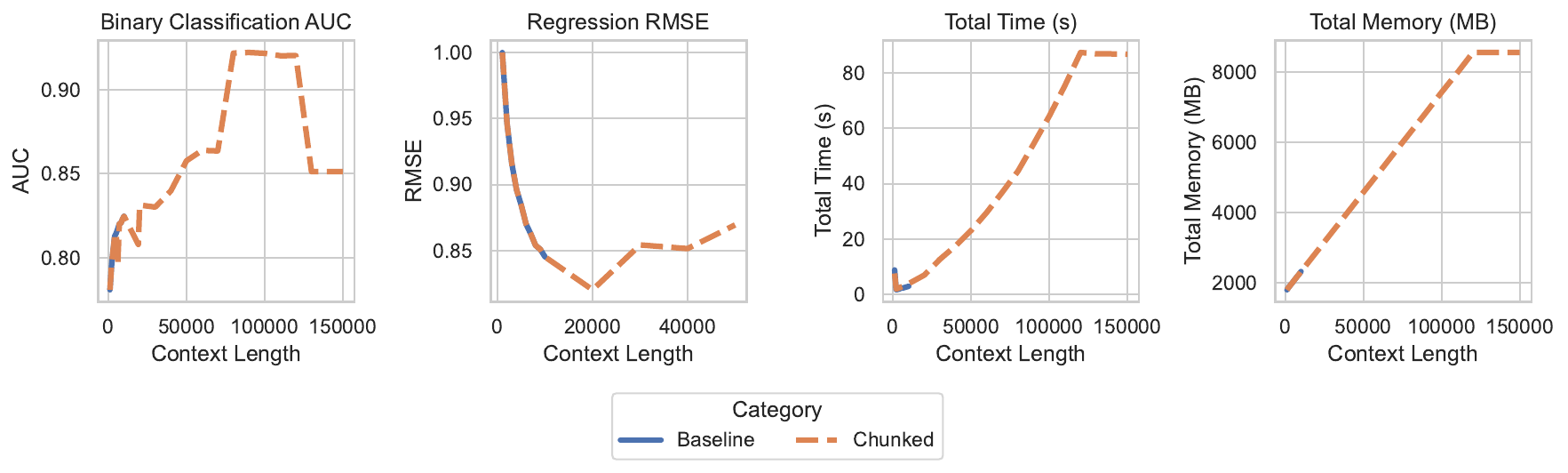}
    \caption{Comparison of our proposed chunked scheme against the standard TabPFN v2 baseline on 15 long-context datasets. For each context length, the training examples are randomly subsampled with a fixed random seed to ensure feasibility for the standard TabPFN. All experiments were conducted on a 32GB NVIDIA V100 GPU.}
    \label{fig:scaling}
\end{figure}

\paragraph{\textbf{Properties.}} Our proposed chunked approach has three main properties:
\begin{enumerate}[label=\arabic*.]
    \item \textbf{Exactness.} Each query position computes its softmax over the full key set; blockwise evaluation is mathematically identical to monolithic attention, up to floating-point associativity.  
    \item \textbf{Memory.} Peak activation memory scales with the chosen chunk size ($\ell$ or $m$), enabling substantially longer contexts without increasing FLOPs.  
    \item \textbf{Compatibility.} The method integrates with any SDPA-compatible backend and preserves permutation invariance along the sample dimension.  
\end{enumerate}
Figure~\ref{fig:scaling} summarizes two key findings discussed in the methodology section. We evaluate on a subset of 15 datasets whose sizes exceed the TabPFN v2 pre-training context length of $10{,}000$ examples by more than $50\%$—datasets that the standard TabPFN cannot handle in full. We refer to these as \emph{long-context} datasets. For each dataset, we progressively subsample the training set up to the maximum size that TabPFN v2 can process. At each context size, we run both the standard TabPFN v2 (baseline) and our chunked scheme, and report averages across all 15 long-context datasets.  

First, we analyze how predictive performance scales with context length. The results show that TabPFN continues to benefit from additional in-context samples even beyond the 10K pre-training limit: average AUC increases steadily up to 100K, while RMSE decreases up to approximately 20K.  

Second, we assess the impact of our chunking scheme. Importantly, no performance degradation is observed relative to the original TabPFN for contexts shorter than 10K. This demonstrates that \textbf{our method preserves predictive accuracy}, in contrast to many tiling-based implementations that often suffer from numerical instabilities or degraded performance.

\section{Experiments}
\label{sec:experiments}
In this section, we evaluate our proposed approach on the recently introduced \emph{TabArena} benchmark \citep{tabarena} and compare it against all methods included in TabArena.  
\subsection{Setup}
\label{subsec:setup}

\paragraph{\textbf{Benchmark.}}
We evaluate our approach on the recently proposed \emph{TabArena} benchmark \citep{tabarena}, which comprises 51 diverse tabular datasets spanning binary and multiclass classification as well as regression tasks.  

From this collection, we select a subset of 15 datasets whose sizes exceed the TabPFN v2 pre-training context length of $10{,}000$ examples by more than $50\%$. We refer to these datasets as \textit{long-context} datasets throughout this work.

\paragraph{\textbf{Task protocol.}}
We follow TabArena’s standardized data splits, preprocessing, and evaluation framework. To ensure fairness and reproducibility, we rely on the officially reported results of all baseline methods. Our approach uses the \textbf{official TabPFN v2 weights without any fine-tuning} and differs only in the computation of attention, where we employ chunked/tiled attention implemented purely in PyTorch.

\paragraph{\textbf{Evaluation metrics.}}
Following TabArena, we report (i) per-dataset metrics (RMSE for regression, AUC for classification) and their aggregates, and (ii) leaderboard-style aggregates (Elo, normalized score, average rank, harmonic-mean rank, \#wins, and improvability).

For the long-context slice, we additionally report \textit{normalized} RMSE and AUC, where the RMSE of each method on a dataset is divided by the worst RMSE among compared methods on that dataset; lower is better for normalized RMSE and higher is better for AUC.

\paragraph{\textbf{Baselines.}}
We compare against strong tree-based and neural baselines used by TabArena, including AutoGluon, CatBoost, LightGBM, XGBoost, TabM, TabDPT, ModernNCA, RealMLP, KNN, and TabICL.
We follow TabArena’s notation: (T) indicates tuned configurations using the benchmark’s search space/budget; (D) indicates out-of-the-box defaults; (E) indicates extra ensembling beyond the default training recipe.

\paragraph{\textbf{Imputation policy.}}
Prior TabPFN v2 numbers on TabArena were partially \emph{imputed} using RandomForest defaults when runs failed or were missing.
In contrast, we \textbf{do not impute} any TabPFNv2 results.
When visualizing historical numbers for comparison, imputed bars are indicated with a stripe pattern (Figure~\ref{fig:tuning-impact-elo}).

\paragraph{\textbf{Compute.}}
All experiments are conducted on single-GPU nodes with at most 24\,GB of VRAM. We report wall-clock training (fit) and inference (predict) times in Tables~\ref{tab:tabarena_full} and \ref{tab:tabarena_long_results_condensed}.

\subsection{Results}
\label{subsec:results}

\begin{figure}[t]
    \centering
    \includegraphics[width=\linewidth]{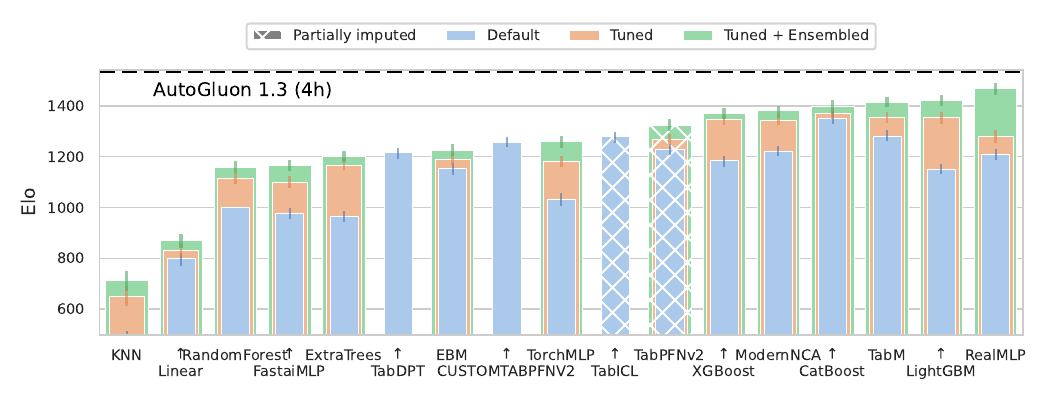}
\caption{\textbf{Leaderboard position and imputation.} Elo trajectories across tuning budgets. 
Unlike prior reports, our \textbf{chunked TabPFN} v2 (\textit{CUSTOMTABPFNV2}) results are reported directly and are \emph{not} imputed; historical imputed entries are indicated with a stripe pattern.}
    \label{fig:tuning-impact-elo}
\end{figure}

\begin{table*}[t]
\centering
\caption{\textbf{TabArena (51 datasets).} Leaderboard-style aggregates following the TabArena protocol. Our \emph{chunked} TabPFN v2 is evaluated in a zero-shot setting (no fine-tuning). Reported times are normalized per 1K examples by the evaluation harness.}

\label{tab:tabarena_full}
\small
\resizebox{\textwidth}{!}{
\input{table_short}
}
\end{table*}

\input{table_longcontext_short} 

\paragraph{\textbf{Full TabArena (51 datasets).}}
Table~\ref{tab:tabarena_full} summarizes performance on the complete benchmark.
Despite operating in a \emph{zero-shot} setting, our \textbf{chunked TabPFN} v2 achieves competitive performance on Elo and normalized score, while also securing multiple dataset-level wins.
The key practical advantage is \textbf{fit-time efficiency}: on average its train $\sim\!$$\mathbf{10^3}\times$ faster than the best-performing AutoGluon (4h) configuration.
Inference time is competitive with deep baselines and faster than AutoGluon, though—as expected—tree ensembles remain the fastest at prediction.

\paragraph{\textbf{Long-context slice (15 datasets).}}
Table~\ref{tab:tabarena_long_results_condensed} reports averages over the long-context datasets. 
Our \textbf{chunked TabPFN} v2 achieves performance comparable to the strongest neural baselines on normalized RMSE and AUC, while being orders of magnitude faster to fit than leading tree-based and AutoML systems (e.g., $\sim$1{,}100$\times$ faster than AutoGluon and $\sim$3{,}000$\times$ faster than tuned CatBoost in average fit time). Moreover, the method is practical on commodity GPUs due to its use of tiled attention.

\paragraph{\textbf{Qualitative takeaways.}}
Across the 15 long-context datasets we consistently observe:
\begin{enumerate}[label=\arabic*.]
    \item \textbf{No degradation with scale.} Increasing the available training context does not harm zero-shot TabPFNv2; performance is flat to improving as $n$ grows.
    \item \textbf{Selective benefits from more data.} Only a subset of datasets show sizeable gains from scaling; others saturate early, suggesting that additional samples are redundant given the model’s prior and the task’s intrinsic difficulty.
    \item \textbf{Robust efficiency.} Chunked attention preserves accuracy while removing the quadratic memory/computation bottleneck, making TabPFN v2 usable on GPUs without FlashAttention support or with tighter VRAM limits.
\end{enumerate}
We provide per-dataset scaling curves and a tile-size ablation in Appendix~\ref{app:scaling}, together with failure cases and an analysis of when more data helps.

\paragraph{\textbf{Summary.}}
Two conclusions follow from the tables and Figure~\ref{fig:tuning-impact-elo}:
\begin{enumerate}[label=\arabic*.]
    \item \textbf{TabPFN v2 is a strong, robust baseline} for tabular learning even without tuning. It trails tuned gradient-boosted trees on leaderboard metrics while being dramatically faster to train.  
    \item \textbf{TabPFN v2 works out-of-the-box for long contexts} (i.e., no post-training is required). For roughly half of the long-context datasets we observe clear gains with more data, while the remainder remain flat, suggesting headroom for data- and task-adaptive selection policies.  
\end{enumerate}

\section{Conclusion}
In this paper, we addressed the long-context bottleneck in the current TabPFN architecture by introducing \textbf{chunked TabPFN}, a simple modification that partitions the attention computation to alleviate memory complexity and avoid out-of-memory errors. To the best of our knowledge, this is the first method to compute TabPFN exactly on long-context datasets without any pre-processing of the training examples or fine-tuning of the model. Our results demonstrate that TabPFN is a strong and robust approach for tabular learning and that its performance extends effectively to long-context settings—an observation that, to our knowledge, has not been previously reported. As future work, we plan to incorporate ring attention mechanisms~\citep{ringattention} to further improve the efficiency and scalability of TabPFN.

\bibliography{main}
\bibliographystyle{tmlr}

\newpage
\appendix

\section{Implementation Details}
\label{app:chunked-attn}

\paragraph{\textbf{Goal.}}
Enable inference with long contexts without altering model parameters or outputs. We keep attention \emph{exact} while reducing peak activation memory by evaluating in tiles.

\paragraph{\textbf{Setup and notation.}}
Let $Q\!\in\!\mathbb{R}^{B\times H\times L_q\times d_k}$, $K,V\!\in\!\mathbb{R}^{B\times H\times L_k\times d_k}$.
We choose a query-tile length $\ell$ and a key/value-tile length $r$ (and optionally a batch tile $m$). We iterate over query tiles $Q^{(c)}$ of length $\ell'\!\le\!\ell$, and for each $Q^{(c)}$, we stream KV tiles $(K^{(t)},V^{(t)})$ of length $r'\!\le\!r$.

\paragraph{\textbf{Core idea (log-sum-exp merge).}}
For a fixed $Q^{(c)}$ we want
\[
\mathrm{Attn}(Q^{(c)},K,V)=\operatorname{softmax}\!\bigl(Z\bigr)V,\quad
Z\!=\!\tfrac{1}{\sqrt{d_k}}\,Q^{(c)}K^\top \in \mathbb{R}^{B\times H\times \ell'\times L_k}.
\]
We never materialize $Z$ in full. Instead, for each KV tile we compute the local logits $Z^{(t)}\!\in\!\mathbb{R}^{B\times H\times \ell'\times r'}$ and update:

\begin{equation}
\begin{split}
\mu' &\leftarrow \max(\mu,\ \max_{r'} Z^{(t)}) \quad \text{(row-wise over keys)},\\
s &\leftarrow s\cdot e^{\mu-\mu'} + \sum_{r'} e^{Z^{(t)}-\mu'},\\
a &\leftarrow a\cdot e^{\mu-\mu'} + \bigl(e^{Z^{(t)}-\mu'}\bigr)V^{(t)},\\
\mu &\leftarrow \mu'.
\end{split}
\end{equation}

At the end, $O^{(c)} = a/s$ equals $\operatorname{softmax}(Z)V$ exactly (modulo floating-point associativity). This is the usual numerically stable log-sum-exp trick applied \emph{incrementally} across KV tiles.

\paragraph{\textbf{Algorithm (pseudocode).}}
We provide a framework-native routine that uses standard tensor ops (matmul/einsum, \texttt{exp}, \texttt{max}, etc.). It supports optional batch tiling when $B$ is large.

\begin{algorithm}[h]
\caption{Exact two-level chunked attention (Q/K/V chunking; framework-native)}
\label{alg:appendix-chunked-attn}
\begin{algorithmic}[1]
\Require $Q\!\in\!\mathbb{R}^{B\times H\times L_q\times d_k}$, $K,V\!\in\!\mathbb{R}^{B\times H\times L_k\times d_k}$; query-tile $\ell$, KV-tile $r$, optional batch-tile $m$
\If{$B>m$} \Comment{Optional: batch tiling}
  \State Split $(Q,K,V)$ into batches of size $\le m$ and process each independently; then concatenate on batch.
\EndIf
\State $\mathcal{O}\gets$ empty list
\For{each query tile $Q^{(c)}$ of length $\ell'\!\le\!\ell$ along $L_q$}
  \State Initialize $\mu\!\leftarrow\!-\infty \in \mathbb{R}^{B\times H\times \ell'\times 1}$, $s\!\leftarrow\!0 \in \mathbb{R}^{B\times H\times \ell'\times 1}$, $a\!\leftarrow\!0 \in \mathbb{R}^{B\times H\times \ell'\times d_k}$
  \For{each KV tile $(K^{(t)},V^{(t)})$ of length $r'\!\le\!r$ along $L_k$}
    \State $Z^{(t)} \leftarrow (Q^{(c)} {K^{(t)}}^\top)/\sqrt{d_k}$ \Comment{$Z^{(t)}\in\mathbb{R}^{B\times H\times \ell'\times r'}$}
    \State $\mu' \leftarrow \max\!\bigl(\mu,\,\max_{r'} Z^{(t)}\bigr)$ \Comment{row-wise over keys}
    \State $s \leftarrow s\cdot e^{\mu-\mu'} + \sum_{r'} e^{Z^{(t)}-\mu'}$
    \State $a \leftarrow a\cdot e^{\mu-\mu'} + \bigl(e^{Z^{(t)}-\mu'}\bigr)V^{(t)}$
    \State $\mu \leftarrow \mu'$
  \EndFor
  \State $O^{(c)} \leftarrow a / s$ \Comment{exact softmax over all keys}
  \State append $O^{(c)}$ to $\mathcal{O}$
\EndFor
\State \Return $\mathrm{concat}(\mathcal{O}\ \text{along } L_q)$
\end{algorithmic}
\end{algorithm}

\paragraph{\textbf{Correctness sketch.}}
Let $z_k$ be logit entries in a row and suppose we split indices into tiles $T_1,\dots,T_T$. The softmax numerator $\sum_k e^{z_k}$ and the weighted sum $\sum_k e^{z_k} v_k$ can be accumulated tile-by-tile using a running max $\mu$:
\[
\sum_k e^{z_k} = \sum_t \sum_{k\in T_t} e^{z_k} = 
\sum_t e^{\mu_t-\mu_T}\!\!\sum_{k\in T_t} e^{z_k-\mu_t},\quad
\mu_T=\max_t \mu_t,\ \mu_t=\max_{k\in T_t} z_k,
\]
which yields the update rules above. The final ratio equals the monolithic softmax output.

\paragraph{\textbf{Complexity.}}
FLOPs match standard attention ($\Theta(BH L_q L_k d_k)$). Peak activation memory depends linearly on $\ell$ and $r$:
\[
\mathcal{O}\!\bigl(BH(\ell r + \ell d_k + r d_k)\bigr)
\]
if block logits are materialized; with fused matmul+reduce this can drop further. Tiling thus trades wall-clock IO and peak memory for negligible loop overhead.

\paragraph{\textbf{Numerical notes.}}
We recommend (i) computing $1/\sqrt{d_k}$ in the working dtype, (ii) keeping the running-$\mu$ and accumulators in the same dtype as logits, and (iii) avoiding dropout inside the accumulation; apply dropout \emph{after} $a/s$ if training-time dropout is required.

\paragraph{\textbf{Implementation hints (PyTorch).}}
Use contiguous tiles shaped as $(B,H,\ell',d_k)$ and $(B,H,r',d_k)$. Prefer \texttt{einsum}/\texttt{matmul} for $Z^{(t)}$; if using \texttt{scaled\_dot\_product\_attention}, you must access logits or per-row LSE to merge tiles—otherwise call SDPA only for the no-chunk case. When KV are shared across fewer heads, broadcast $K,V$ across heads once per outer loop. Causal masks (if any) restrict valid $r'$ per tile; the recurrence is unchanged.

\paragraph{\textbf{When to chunk $B$.}}
If $B$ is too large for memory, first tile batches by $m$ (outermost loop), then apply the $Q$/$KV$ tiling inside each batch tile. This preserves exactness and amortizes kernel launches.

\paragraph{\textbf{Limitations.}}
Tiling increases the number of small GEMMs; on some backends this can slightly reduce throughput. We found the reduction in peak memory and the ability to run much longer contexts outweigh this overhead in long-context regimes.

\section{Full Results on TabArena}
\label{app:full-results}

\begin{figure}[t]
    \centering
    \includegraphics[width=\linewidth]{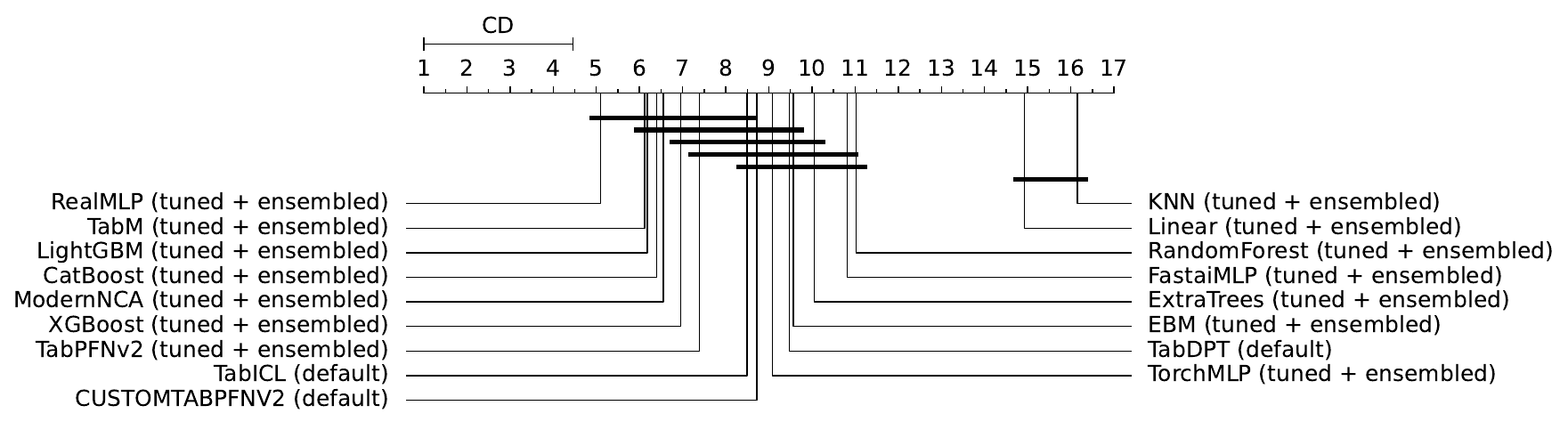}
    \caption{Critical difference diagram for tuned and ensembled methods on the full benchmark. Horizontal bars connect methods that are not statistically significantly different. Our approach (\textit{CUSTOMTABPFN2}) achieves statistically comparable results to tree-based ensembled methods.}
    \label{fig:critical-diagram}
\end{figure}

We provide full results on the TabArena benchmark in Table~\ref{tab:full_results}. The critical diagram is presented in Figure~\ref{fig:critical-diagram}. Crucially, when looking at the figures, unlike for the original TabArena, we did not impute the results of TabPFN.

\begin{table*}[t]
\centering
\caption{Full results on \textbf{TabArena (51 datasets)}. Leaderboard-style aggregates copied from the TabArena protocol. Our \textbf{chunked TabPFN} is evaluated zero-shot (no fine-tuning). Times are normalized per 1K examples by the harness.}
\label{tab:full_results}
\resizebox{\textwidth}{!}{
\input{table}
}
\end{table*}

\section{Per-dataset Analysis on TabArena Long Datasets}
\label{app:scaling}

\input{table_longcontext}

We provide the per-dataset plots of error, computation time, and memory scaling for the 15 long-context datasets described in Table~\ref{tab:tabarena_long_results} selected from TabArena \cite{tabarena}. We see that for most datasets the error / metric continues to improve well past the pre-training context length of 10,000. In general, we observe 3 types of behavior: 1) early plateua: model stops improving but does not degrade beyond certain length happens on (i)amazon\_employee, (ii)diabetes, (iii)bank marketing, (iv)customer\_satisafaction, (v)food\_delivery, (vi)kdd\_cup, (vii)sdss17 ( 7 out of 15 datasets) 2) continuous improvement: model continues improving on (i) superconductivty, (ii) physiochemical\_protein, (iii) hr\_analytics, (iv) houses, (v) apsfailure, (vi) diamonds, (vii)credit\_card (7 out of 15 datasets); 3) model failure: on 1 dataset (givemesomecredit) the model results decrease with more context -- although they stay within a narrow band of AUC values. 

It is an interesting direction to study the underlying causes of the above observed behavior. We hypoethsize that a combination of 1) dataset hardness 2) model priors could be playing a role here. 

\begin{figure}[t]
    \centering
    \includegraphics[width=0.9\linewidth]{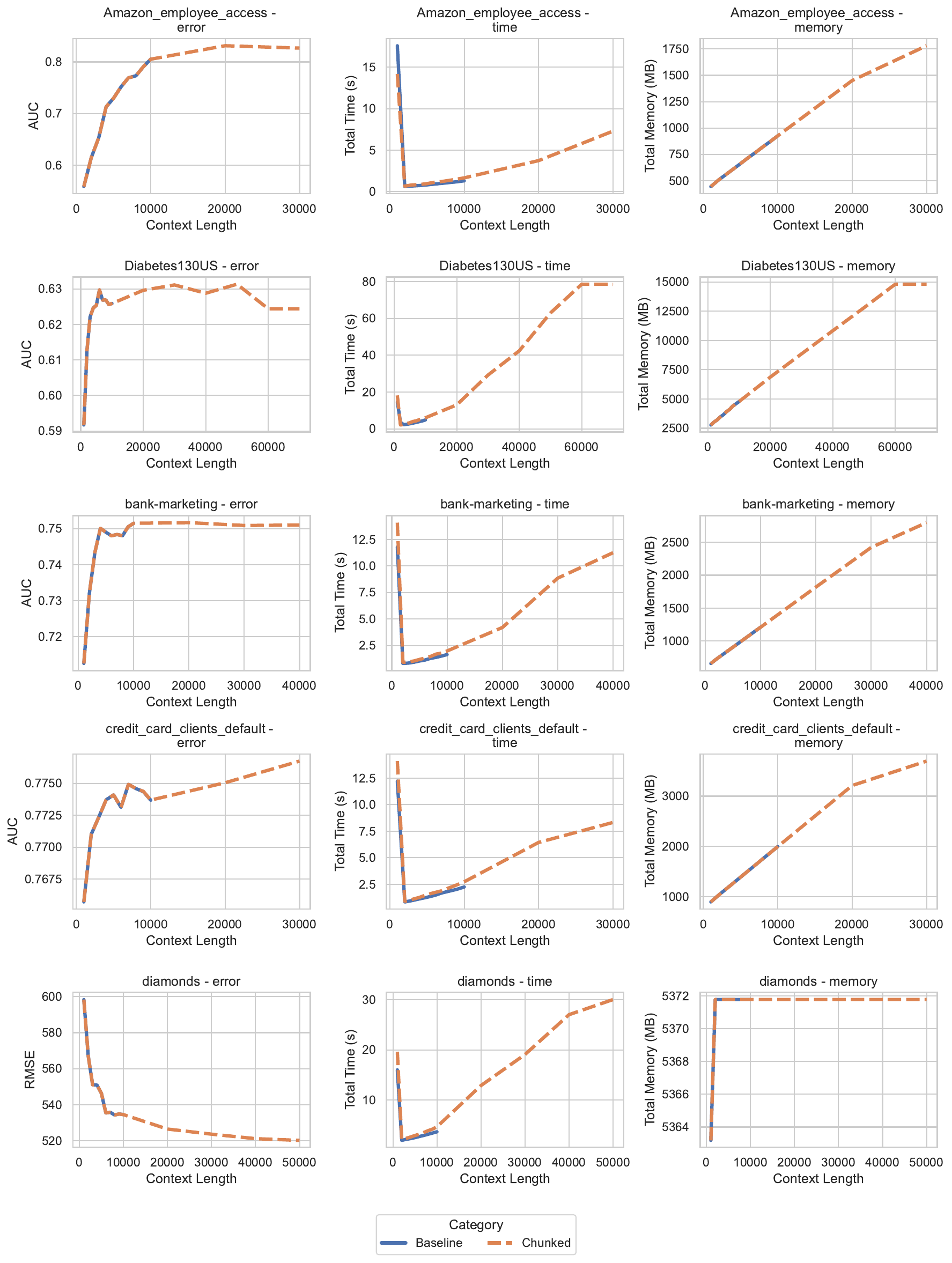}
    \caption[Scaling vs.\ context length (part I/III)]{\textbf{Scaling vs.\ context length (part I/III).}
  Error (RMSE~$(\downarrow)$~/~AUC~$\uparrow$/Accuracy$(\uparrow)$), wall-clock time (s), and peak GPU memory (MB) for \emph{TabPFNv2 (baseline)} vs.\ \emph{TabPFNv2 (chunked)}.
  Datasets 1–5 of the 15 long-context tasks.}
    \label{fig:scaling_dataset1-5}
\end{figure}

\begin{figure}[t]
    \centering
    \includegraphics[width=0.9\linewidth]{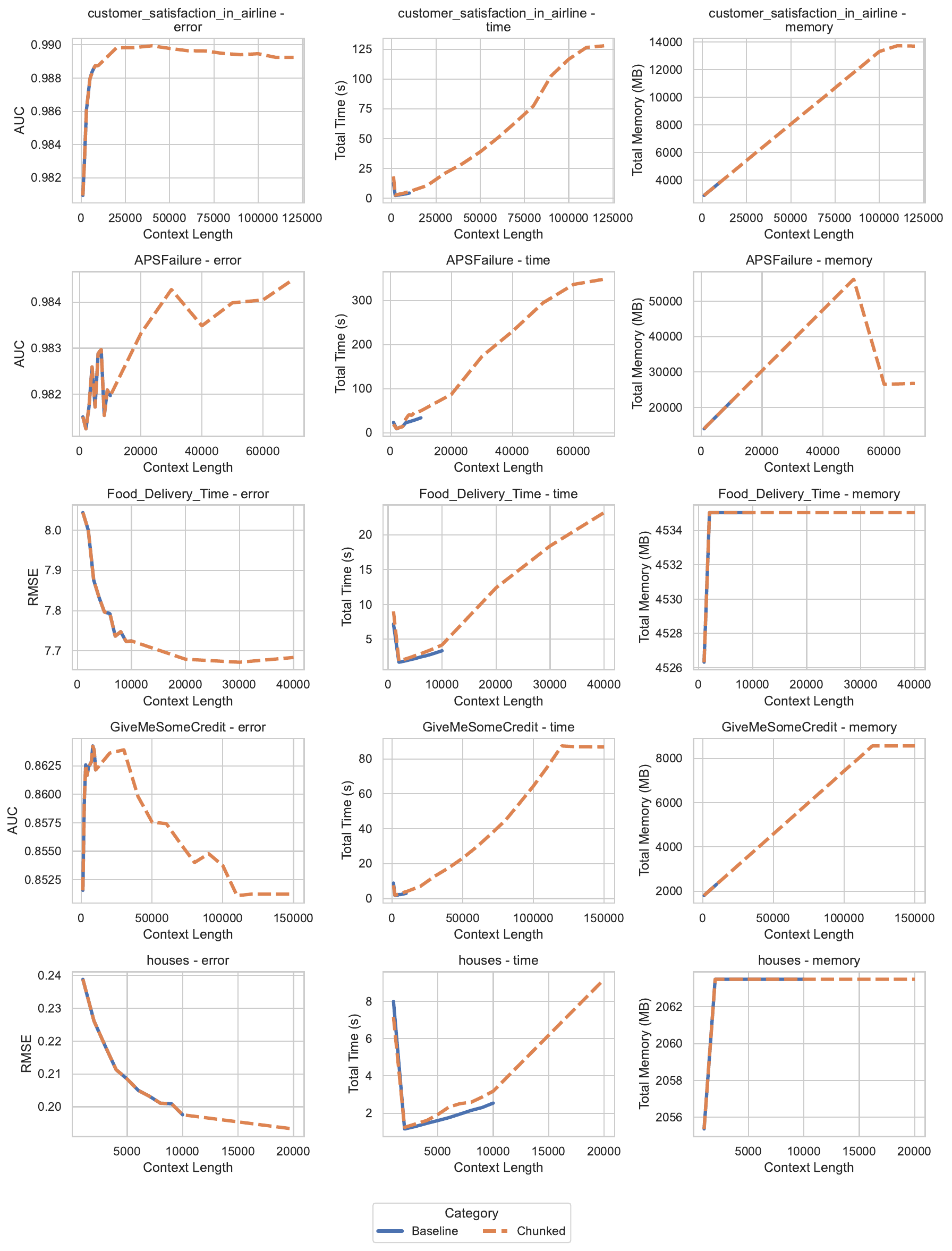}
    \caption[Scaling vs.\ context length (part II/III)]{\textbf{Scaling vs.\ context length (part II/III).}
  Error (RMSE~$(\downarrow)$~/~AUC~$\uparrow$/Accuracy$(\uparrow)$), wall-clock time (s), and peak GPU memory (MB) for \emph{TabPFNv2 (baseline)} vs.\ \emph{TabPFNv2 (chunked)}.
  Datasets 6–10 of the 15 long-context tasks.}
    \label{fig:scaling_dataset6-10}
\end{figure}

\begin{figure}[t]
    \centering
    \includegraphics[width=0.9\linewidth]{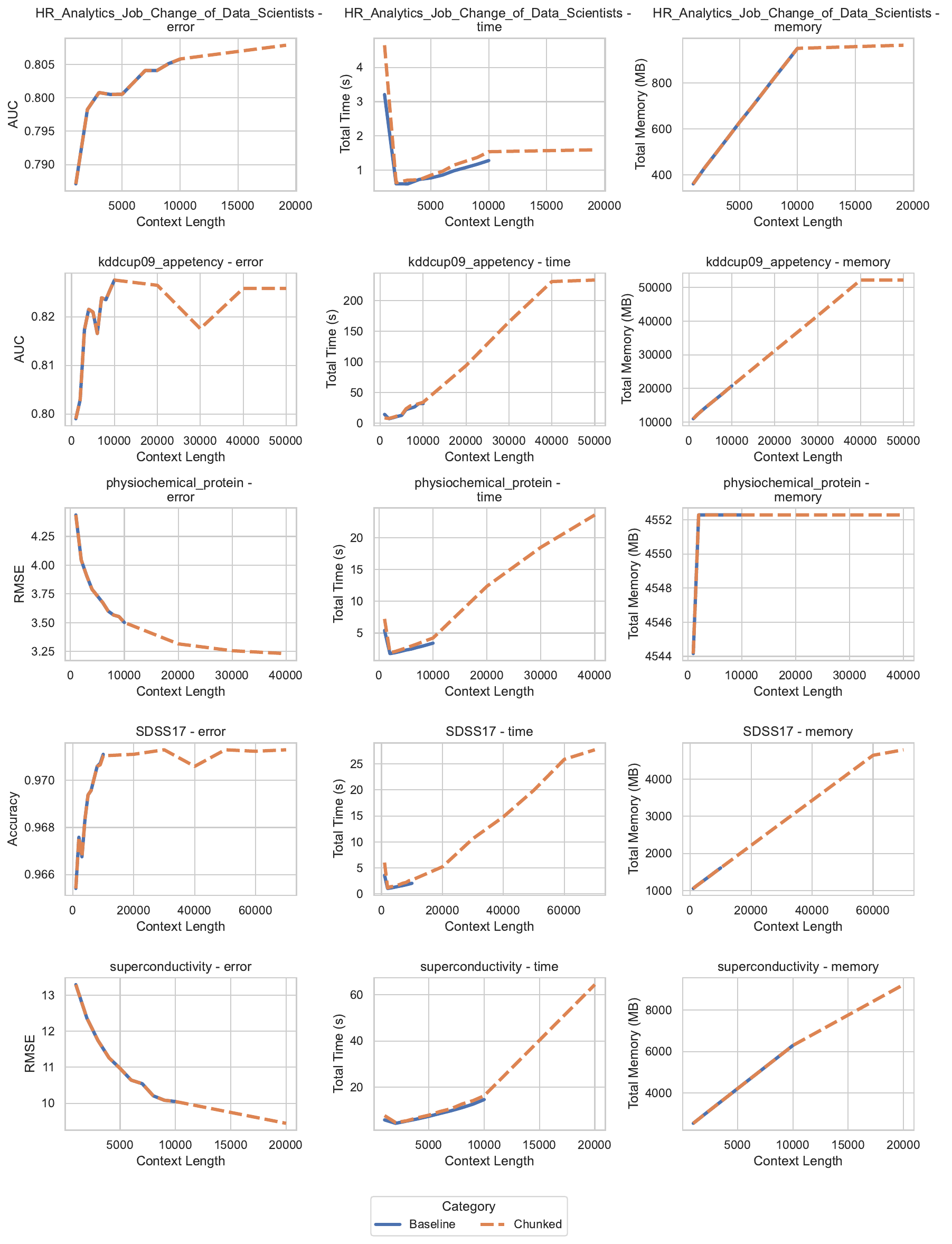}
    \caption[Scaling vs.\ context length (part III/III)]{\textbf{Scaling vs.\ context length (part II/III).}
  Error (RMSE~$(\downarrow)$~/~AUC~$\uparrow$/Accuracy$(\uparrow)$), wall-clock time (s), and peak GPU memory (MB) for \emph{TabPFNv2 (baseline)} vs.\ \emph{TabPFNv2 (chunked)}.
  Datasets 11–15 of the 15 long-context tasks.}
    \label{fig:scaling_dataset11-15}
\end{figure}

\end{document}

%% file: table_short.tex
\begin{tabular}{llcccccrr}
\toprule
\textbf{Model} & \textbf{Elo ($\uparrow$)} & \textbf{Norm.} & \textbf{Avg.} & \textbf{Harm.} & \textbf{\#wins ($\uparrow$)} & \textbf{Improva-} & \textbf{Train time} & \textbf{Predict time} \\
 &  & \textbf{score ($\uparrow$)} & \textbf{rank ($\downarrow$)} & \textbf{mean} &  & \textbf{bility ($\downarrow$)} & \textbf{per 1K [s]} & \textbf{per 1K [s]} \\
 &  &  &  & \textbf{rank ($\downarrow$)} &  &  &  &  \\
\midrule
AutoGluon 1.3 (4h) & \textcolor{gold}{\textbf{1536${}_{-27,+26}$}} & \textcolor{gold}{\textbf{0.604}} & \textcolor{gold}{\textbf{7.9}} & \textcolor{gold}{\textbf{3.3}} & \textcolor{silver}{\textbf{7}} & \textcolor{gold}{\textbf{6.4\%}} & 1453.27 & 3.15 \\
CatBoost (T) & 1372${}_{-25,+21}$ & 0.370 & 14.5 & 8.7 & 0 & 10.0\% & 1665.53 & 0.07 \\
TabM (T) & 1354${}_{-23,+22}$ & 0.369 & 15.5 & 8.5 & 0 & 9.9\% & 3133.91 & 0.13 \\
LightGBM (T) & 1353${}_{-25,+24}$ & 0.322 & 15.4 & 10.6 & 0 & 11.1\% & 416.56 & 0.38 \\
XGBoost (T) & 1346${}_{-20,+19}$ & 0.312 & 15.8 & 9.0 & 1 & 11.0\% & 700.96 & 0.21 \\
RealMLP (T) & 1279${}_{-25,+25}$ & 0.247 & 19.1 & 13.3 & 0 & 12.3\% & 6559.12 & 0.35 \\
TabICL (D) & 1278${}_{-24,+19}$ & 0.299 & 19.3 & 6.2 & \textcolor{bronze}{\textbf{4}} & 11.6\% & 6.86 & 1.52 \\
TabPFNv2 (T) & 1268${}_{-21,+24}$ & 0.303 & 19.8 & 6.8 & 1 & 12.1\% & 2942.08 & 0.26 \\
CustomTabPFNv2 (D) & 1254${}_{-17,+24}$ & 0.308 & 20.4 & 6.3 & 3 & 12.3\% & 8.67 & 4.45 \\
TabPFNv2 (D) & 1227${}_{-22,+22}$ & 0.262 & 22.0 & 7.8 & 1 & 13.1\% & 3.27 & 0.32 \\
XGBoost (D) & 1184${}_{-25,+19}$ & 0.139 & 24.3 & 14.3 & 1 & 14.3\% & 2.06 & 0.12 \\
KNN (D) & 475${}_{-43,+38}$ & 0.000 & 44.6 & 44.2 & 0 & 54.7\% & 0.05 & 0.02 \\
\bottomrule
\end{tabular}

%% file: table_longcontext_short.tex

\begin{table}[t]
  \centering
  \caption{\textbf{Long-context (15 datasets).} Abridged comparison of our \textbf{chunked TabPFN} v2 vs.\ representative baselines using normalized metrics (lower is better for RMSE; higher is better for AUC). Our method attains competitive accuracy while being orders-of-magnitude faster to \emph{fit}. (T) tuned, (D) default, (E) extra ensembling.}
  \label{tab:tabarena_long_results_condensed}
    \resizebox{\textwidth}{!}{\begin{tabular}{lrrrr}
    \toprule
    Method & Avg. norm. RMSE ($\downarrow$) & Avg. AUC ($\uparrow$) & Avg. Fit Time (s) & Avg. Eval Time (s) \\
    \midrule
    AutoGluon & 0.523 & 0.846 & 13012.001 & 100.151 \\
    TabDPT (D) & 0.524 & 0.818 & 376.869 & 158.585 \\
    ModernNCA (D) & 0.534 & 0.833 & 777.194 & 23.230 \\
    TabPFNv2 (chunked) &{0.534} &{0.826} &{11.524} &{63.748} \\
    CatBoost (T+E) & 0.538 & 0.845 & 34537.203 & 6.370 \\
    CatBoost (T) & 0.538 & 0.845 & 34537.203 & 0.779 \\
    TabM (D) & 0.551 & 0.833 & 331.131 & 2.096 \\
    CatBoost (D) & 0.543 & 0.845 & 262.200 & 0.456 \\
    KNN (D) & 0.919 & 0.643 & 1.605 & 0.539 \\
    \bottomrule
    \end{tabular}
    }
\end{table}

%% file: table.tex
\begin{tabular}{llcccccrr}
\toprule
\textbf{Model} & \textbf{Elo ($\uparrow$)} & \textbf{Norm.} & \textbf{Avg.} & \textbf{Harm.} & \textbf{\#wins ($\uparrow$)} & \textbf{Improva-} & \textbf{Train time} & \textbf{Predict time} \\
 &  & \textbf{score ($\uparrow$)} & \textbf{rank ($\downarrow$)} & \textbf{mean} &  & \textbf{bility ($\downarrow$)} & \textbf{per 1K [s]} & \textbf{per 1K [s]} \\
 &  &  &  & \textbf{rank ($\downarrow$)} &  &  &  &  \\
\midrule
AutoGluon 1.3 (4h) & \textcolor{gold}{\textbf{1536${}_{-27,+26}$}} & \textcolor{gold}{\textbf{0.604}} & \textcolor{gold}{\textbf{7.9}} & \textcolor{gold}{\textbf{3.3}} & \textcolor{silver}{\textbf{7}} & \textcolor{gold}{\textbf{6.4\%}} & 1453.27 & 3.15 \\
RealMLP (T+E) & \textcolor{silver}{\textbf{1468${}_{-26,+23}$}} & \textcolor{silver}{\textbf{0.523}} & \textcolor{silver}{\textbf{10.3}} & 5.0 & 3 & \textcolor{silver}{\textbf{8.3\%}} & 6559.12 & 8.60 \\
LightGBM (T+E) & \textcolor{bronze}{\textbf{1422${}_{-24,+20}$}} & 0.399 & \textcolor{bronze}{\textbf{12.3}} & 8.5 & 0 & 10.2\% & 416.56 & 2.24 \\
TabM (T+E) & 1414${}_{-21,+22}$ & 0.444 & 12.6 & 6.4 & 2 & \textcolor{bronze}{\textbf{8.9\%}} & 3133.91 & 1.27 \\
CatBoost (T+E) & 1396${}_{-29,+26}$ & 0.402 & 13.4 & 8.7 & 0 & 9.5\% & 1665.53 & 0.56 \\
ModernNCA (T+E) & 1380${}_{-27,+21}$ & \textcolor{bronze}{\textbf{0.447}} & 14.1 & \textcolor{bronze}{\textbf{4.8}} & \textcolor{bronze}{\textbf{4}} & 9.9\% & 4618.50 & 7.74 \\
CatBoost (T) & 1372${}_{-25,+21}$ & 0.370 & 14.5 & 8.7 & 0 & 10.0\% & 1665.53 & 0.07 \\
XGBoost (T+E) & 1371${}_{-25,+21}$ & 0.342 & 14.6 & 8.3 & 1 & 10.8\% & 700.96 & 1.44 \\
TabM (T) & 1354${}_{-23,+22}$ & 0.369 & 15.5 & 8.5 & 0 & 9.9\% & 3133.91 & 0.13 \\
LightGBM (T) & 1353${}_{-25,+24}$ & 0.322 & 15.4 & 10.6 & 0 & 11.1\% & 416.56 & 0.38 \\
CatBoost (D) & 1350${}_{-23,+25}$ & 0.317 & 15.6 & 9.5 & 1 & 10.8\% & 6.70 & 0.09 \\
XGBoost (T) & 1346${}_{-20,+19}$ & 0.312 & 15.8 & 9.0 & 1 & 11.0\% & 700.96 & 0.21 \\
ModernNCA (T) & 1343${}_{-21,+25}$ & 0.361 & 15.9 & 7.0 & 2 & 10.5\% & 4618.50 & 0.47 \\
TabPFNv2 (T+E) & 1324${}_{-25,+26}$ & 0.423 & 17.0 & \textcolor{silver}{\textbf{3.5}} & \textcolor{gold}{\textbf{11}} & 10.4\% & 2942.08 & 17.37 \\
TabM (D) & 1280${}_{-17,+27}$ & 0.266 & 19.1 & 12.1 & 0 & 12.7\% & 11.56 & 0.13 \\
RealMLP (T) & 1279${}_{-25,+25}$ & 0.247 & 19.1 & 13.3 & 0 & 12.3\% & 6559.12 & 0.35 \\
TabICL (D) & 1278${}_{-24,+19}$ & 0.299 & 19.3 & 6.2 & \textcolor{bronze}{\textbf{4}} & 11.6\% & 6.86 & 1.52 \\
TabPFNv2 (T) & 1268${}_{-21,+24}$ & 0.303 & 19.8 & 6.8 & 1 & 12.1\% & 2942.08 & 0.26 \\
TorchMLP (T+E) & 1259${}_{-27,+22}$ & 0.193 & 20.3 & 14.2 & 0 & 12.4\% & 2832.80 & 1.80 \\
CUSTOMTABPFNV2 (D) & 1254${}_{-17,+24}$ & 0.308 & 20.4 & 6.3 & 3 & 12.3\% & 8.67 & 4.45 \\
TabPFNv2 (D) & 1227${}_{-22,+22}$ & 0.262 & 22.0 & 7.8 & 1 & 13.1\% & 3.27 & 0.32 \\
EBM (T+E) & 1225${}_{-22,+24}$ & 0.203 & 21.9 & 12.7 & 0 & 15.8\% & 1323.39 & 0.18 \\
ModernNCA (D) & 1221${}_{-21,+22}$ & 0.179 & 22.2 & 11.8 & 1 & 15.4\% & 13.74 & 0.32 \\
TabDPT (D) & 1215${}_{-23,+20}$ & 0.265 & 22.6 & 6.5 & \textcolor{bronze}{\textbf{4}} & 14.8\% & 20.56 & 8.62 \\
RealMLP (D) & 1207${}_{-20,+22}$ & 0.138 & 23.0 & 16.7 & 0 & 14.1\% & 21.59 & 1.49 \\
ExtraTrees (T+E) & 1200${}_{-22,+24}$ & 0.159 & 23.3 & 14.1 & 0 & 15.8\% & 191.44 & 0.76 \\
EBM (T) & 1190${}_{-24,+23}$ & 0.168 & 24.0 & 12.1 & 1 & 16.5\% & 1323.39 & 0.02 \\
XGBoost (D) & 1184${}_{-25,+19}$ & 0.139 & 24.3 & 14.3 & 1 & 14.3\% & 2.06 & 0.12 \\
TorchMLP (T) & 1183${}_{-23,+21}$ & 0.149 & 24.2 & 18.1 & 0 & 14.3\% & 2832.80 & 0.11 \\
ExtraTrees (T) & 1168${}_{-22,+24}$ & 0.150 & 25.0 & 13.4 & 0 & 16.7\% & 191.44 & 0.10 \\
FastaiMLP (T+E) & 1165${}_{-22,+23}$ & 0.133 & 25.2 & 16.7 & 0 & 16.5\% & 594.95 & 4.65 \\
RandomForest (T+E) & 1158${}_{-25,+23}$ & 0.115 & 25.5 & 16.0 & 0 & 16.4\% & 377.08 & 0.75 \\
EBM (D) & 1153${}_{-24,+23}$ & 0.148 & 25.9 & 13.8 & 1 & 17.4\% & 5.48 & 0.06 \\
LightGBM (D) & 1152${}_{-22,+20}$ & 0.106 & 26.0 & 22.0 & 0 & 15.3\% & 2.20 & 0.17 \\
RandomForest (T) & 1113${}_{-20,+22}$ & 0.072 & 27.8 & 22.3 & 0 & 17.3\% & 377.08 & 0.09 \\
FastaiMLP (T) & 1100${}_{-23,+24}$ & 0.088 & 28.5 & 19.9 & 0 & 18.1\% & 594.95 & 0.34 \\
TorchMLP (D) & 1031${}_{-24,+25}$ & 0.039 & 31.7 & 28.1 & 0 & 19.8\% & 8.96 & 0.13 \\
RandomForest (D) & 1000${}_{-0,+0}$ & 0.020 & 32.9 & 27.8 & 0 & 22.6\% & 0.43 & 0.05 \\
FastaiMLP (D) & 976${}_{-24,+21}$ & 0.038 & 33.9 & 30.6 & 0 & 22.2\% & 3.12 & 0.31 \\
ExtraTrees (D) & 967${}_{-26,+21}$ & 0.032 & 34.2 & 30.2 & 0 & 24.4\% & 0.26 & 0.05 \\
Linear (T+E) & 870${}_{-29,+26}$ & 0.022 & 37.5 & 21.6 & 1 & 32.7\% & 47.11 & 0.16 \\
Linear (T) & 832${}_{-25,+25}$ & 0.015 & 38.6 & 31.2 & 0 & 33.3\% & 47.11 & 0.06 \\
Linear (D) & 800${}_{-28,+24}$ & 0.010 & 39.5 & 37.4 & 0 & 34.6\% & 1.53 & 0.09 \\
KNN (T+E) & 713${}_{-40,+38}$ & 0.010 & 41.3 & 37.0 & 0 & 45.8\% & 2.61 & 0.16 \\
KNN (T) & 653${}_{-39,+39}$ & 0.012 & 42.4 & 34.0 & 0 & 47.5\% & 2.61 & 0.03 \\
KNN (D) & 475${}_{-43,+38}$ & 0.000 & 44.6 & 44.2 & 0 & 54.7\% & 0.05 & 0.02 \\
\bottomrule
\end{tabular}

%% file: table_longcontext.tex

\begin{table}[htb]
  \centering
  \caption{Performance of our \textbf{chunked TabPFN} v2 on long-context TabArena datasets (context length 10,000) compared to other methods. For regression tasks, RMSE is normalized by the maximum RMSE observed for that dataset across all methods. (T) denotes tuned, (D) denotes default, and (E) denotes extra ensembling.}

  \label{tab:tabarena_long_results}
   \resizebox{\textwidth}{!}{
   \begin{tabular}{lrrrr}
\toprule
Method & Avg. norm. RMSE ($\downarrow$) & Avg. AUC error, $\uparrow$) & Avg. Fit Time (s) & Avg. Eval Time (s) \\
\midrule
Portfolio & 0.512 & 0.846 & 13385.394 & 139.272 \\
ModernNCA (T+E) & 0.518 & 0.839 & 134211.484 & 1337.717 \\
AutoGluon & 0.523 & 0.846 & 13012.001 & 100.151 \\
TabDPT (D) & 0.524 & 0.818 & 376.869 & 158.585 \\
RealMLP (T+E) & 0.525 & 0.841 & 216864.525 & 108.707 \\
ModernNCA (T) & 0.527 & 0.838 & 134211.484 & 48.345 \\
ModernNCA (D) & 0.534 & 0.833 & 777.194 & 23.230 \\
TabPFNv2 (chunked) & 0.534 & 0.826 & 11.524 & 63.748 \\
TabM (T+E) & 0.535 & 0.838 & 88971.102 & 18.308 \\
LightGBM (T+E) & 0.537 & 0.841 & 6004.428 & 60.489 \\
RealMLP (T) & 0.538 & 0.837 & 216864.525 & 4.208 \\
CatBoost (T+E) & 0.538 & 0.845 & 34537.203 & 6.370 \\
LightGBM (T) & 0.538 & 0.838 & 6004.428 & 9.289 \\
CatBoost (T) & 0.538 & 0.845 & 34537.203 & 0.779 \\
TabM (T) & 0.539 & 0.837 & 88971.102 & 2.006 \\
XGBoost (T+E) & 0.541 & 0.842 & 11910.794 & 17.903 \\
XGBoost (T) & 0.541 & 0.841 & 11910.794 & 3.905 \\
CatBoost (D) & 0.543 & 0.845 & 262.200 & 0.456 \\
XGBoost (D) & 0.548 & 0.836 & 25.601 & 1.573 \\
TorchMLP (T+E) & 0.548 & 0.839 & 74418.922 & 34.416 \\
TabM (D) & 0.551 & 0.833 & 331.131 & 2.096 \\
LightGBM (D) & 0.551 & 0.832 & 12.456 & 2.786 \\
TorchMLP (T) & 0.553 & 0.836 & 74418.922 & 1.889 \\
RF (T+E) & 0.559 & 0.835 & 4462.099 & 5.294 \\
RF (T) & 0.560 & 0.834 & 4462.099 & 0.527 \\
RealMLP (D) & 0.562 & 0.836 & 756.178 & 3.567 \\
ExtraTrees (T+E) & 0.562 & 0.833 & 1666.556 & 6.465 \\
ExtraTrees (T) & 0.563 & 0.832 & 1666.556 & 0.686 \\
TabICL (D) & 0.570 & 0.836 & 245.148 & 45.344 \\
RF (D) & 0.570 & 0.817 & 11.507 & 0.436 \\
TorchMLP (D) & 0.575 & 0.832 & 172.946 & 1.255 \\
ExtraTrees (D) & 0.577 & 0.813 & 4.669 & 0.496 \\
FastAIMLP (T+E) & 0.591 & 0.835 & 14601.545 & 31.424 \\
FastAIMLP (T) & 0.593 & 0.831 & 14601.545 & 1.607 \\
EBM (T+E) & 0.595 & 0.837 & 86608.087 & 2.393 \\
EBM (T) & 0.598 & 0.836 & 86608.087 & 0.268 \\
EBM (D) & 0.601 & 0.834 & 327.139 & 0.272 \\
FastAIMLP (D) & 0.621 & 0.822 & 61.772 & 2.081 \\
Linear (T+E) & 0.828 & 0.820 & 416.370 & 3.427 \\
Linear (T) & 0.833 & 0.819 & 416.370 & 0.940 \\
KNN (T+E) & 0.846 & 0.695 & 403.372 & 32.053 \\
KNN (T) & 0.853 & 0.692 & 403.372 & 4.805 \\
Linear (D) & 0.901 & 0.819 & 8.487 & 0.999 \\
KNN (D) & 0.919 & 0.643 & 1.605 & 0.539 \\
\bottomrule
\end{tabular}
   }
\end{table}